\documentclass[journal, twoside, web]{ieeecolor}
\usepackage{arxiv}
\usepackage{cite}
\usepackage{amsmath,amssymb,amsfonts}
\usepackage{algorithmic}
\usepackage{graphicx}
\usepackage{textcomp}
\usepackage{multirow}
\usepackage{color}
\usepackage{caption}
\usepackage{subcaption}
\usepackage{array}
\usepackage{ifthen}
\usepackage[normalem]{ulem}

\newboolean{image_metrics}
\newboolean{video_metrics}
\definecolor{teal}{RGB}{0, 128, 128}
\definecolor{highlight}{RGB}{0, 0, 0}

\newcolumntype{L}{>{\arraybackslash}m{0.7\textwidth}}

\def\BibTeX{{\rm B\kern-.05em{\sc i\kern-.025em b}\kern-.08em
    T\kern-.1667em\lower.7ex\hbox{E}\kern-.125emX}}
\begin{document}

\title{\textcolor{black}{Weakly Supervised Temporal Convolutional Networks for Fine-grained Surgical Activity Recognition}}
\author{Sanat Ramesh,
Diego Dall'Alba,
Cristians Gonzalez,
Tong Yu,
Pietro Mascagni,
Didier Mutter,
Jacques Marescaux, 
Paolo Fiorini, and
Nicolas Padoy
\thanks{This work has received funding from the European Union's Horizon 2020 research and innovation programme under the Marie Sklodowska-Curie grant agreement No 813782 - project ATLAS. This work was also supported by French state funds managed within the Investissements d'Avenir program by BPI France (project CONDOR) and by the ANR (ANR-16-CE33-0009, ANR-10-IAHU-02).}
\thanks{Sanat Ramesh, Diego Dall'Alba and Paolo Fiorini are with
              Altair Robotics Lab, Department of Computer Science, University of Verona, Verona, Italy
             (email: \{sanat.ramesh, diego.dallalba, paolo.fiorini\}@univr.it)}
\thanks{Sanat Ramesh, Tong Yu, Pietro Mascagni and Nicolas Padoy are with
              ICube, University of Strasbourg, CNRS, IHU Strasbourg, France \\
             (email: \{tyu, p.mascagni, npadoy\}@unistra.fr)}
\thanks{Cristians Gonzalez and Didier Mutter are with
              University Hospital of Strasbourg, IHU Strasbourg, France}
\thanks{Didier Mutter and Jacques Marescaux are with IRCAD, France}
\thanks{Pietro Mascagni is with 
              Fondazione Policlinico Universitario Agostino Gemelli IRCCS, Rome, Italy}
\thanks{Copyright © 2023 IEEE. Personal use of this material is permitted. Permission from IEEE must be obtained for all other uses, in any current or future media, including reprinting/republishing this material for advertising or promotional purposes, creating new collective works, for resale or redistribution to servers or lists, or reuse of any copyrighted component of this work in other works.}
}
\maketitle

\setboolean{video_metrics}{true}

\begin{abstract}
Automatic recognition of fine-grained surgical activities, called steps, is a challenging but crucial task for intelligent intra-operative computer assistance. The development of current vision-based activity recognition methods relies heavily on a high volume of manually annotated data. This data is difficult and time-consuming to generate and requires domain-specific knowledge. In this work, we propose to use coarser and easier-to-annotate activity labels, namely phases, as weak supervision to learn step recognition with fewer step annotated videos. We introduce a step-phase dependency loss to exploit the weak supervision signal. We then employ a Single-Stage Temporal Convolutional Network (SS-TCN) with a ResNet-50 backbone, trained in an end-to-end fashion from weakly annotated videos, for temporal activity segmentation and recognition. We extensively evaluate and show the effectiveness of the proposed method on a large video dataset consisting of 40 laparoscopic gastric bypass procedures and the public benchmark CATARACTS containing 50 cataract surgeries.
\end{abstract}


\section{Introduction}
\label{sec:introduction}
\IEEEPARstart{R}{esearch} in developing advanced clinical decision support systems in computer-assisted interventions (CAI) and robot-assisted surgeries (RAS) for the demanding situations of a modern Operating Room (OR) \cite{Cleary2005495, MaierHein2017, 8880624} has seen significant progress in the last decade. One of the primary functions of these advanced systems is automatic surgical workflow analysis, i.e., reliable recognition of the current surgical activities. Surgical activity recognition could play a key role in assisting clinical decisions, report generation, and data annotation by providing valuable semantic information. 

Depending on the level of granularity, a surgical procedure can be decomposed into activities, such as the whole procedure, phases, stages, steps, and actions \cite{Katic2015, SAGESMeireles2021}. 
Surgical phases are defined as a set of fundamental surgical aims to accomplish in order to successfully complete the surgical procedure. Similarly, steps are defined as a set of surgical actions to perform in order to accomplish a surgical phase. 
These definitions help clinicians define an ontology for each procedure, e.g. \cite{Charrire2017, Ramesh2021} define ontologies for cataract and gastric bypass procedures. Although the ontologies are well defined, automatically recognizing these activities from available endoscopic videos is a topic of high interest.

Phase recognition has received a lot of attention and is a very active area of research in the medical computer vision community \cite{Twinanda2017, Zisimopoulos2018DeepPhaseSP, Jin2018, Jin2020MultiTaskRC, Czempiel2020TeCNOSP}. Alongside phases, there has been substantial research focusing on fine-grained activities such as robotic gestures
\cite{Zappella2013, LeaGestures2015, LeaGestures2016, Ahmidi2017,10.1007/978-3-030-00937-3_29, FunkeBOBWS19, Gao2020AutomaticGR},
action triplets \cite{10.1007/978-3-030-59716-0_35}, and instrument detection and tracking \cite{AlHajj2018, Nwoye2019WeaklySC, Jin2020MultiTaskRC}. Recently, there has been a surge of research works focusing particularly on step recognition \cite{Charriere2014, Charrire2017, Ramesh2021}. 

While steps define a surgical workflow at a more fine-grained level than phases, the time required to annotate a dataset with steps is significantly higher than with phase annotations. For example, in Laparoscopic Roux-en-Y gastric bypass (LRYGB) procedures, the workflow consists of 44 steps and 11 phases (Table \ref{tab2:lrygb_phst}). Precisely defining and annotating all the steps requires a considerably higher time of experts due to the number of steps and more importantly lower inter-class variances between steps. Since recent works in surgical phase/step recognition employ deep learning models, they rely on the availability of large-scale annotated datasets. Curation of these annotated datasets is difficult and time-consuming as these tasks require domain-specific medical knowledge.

\begin{figure*}[!t]
\centering
\begin{subfigure}[b]{1\textwidth}
    \centering
    \includegraphics[width=\textwidth]{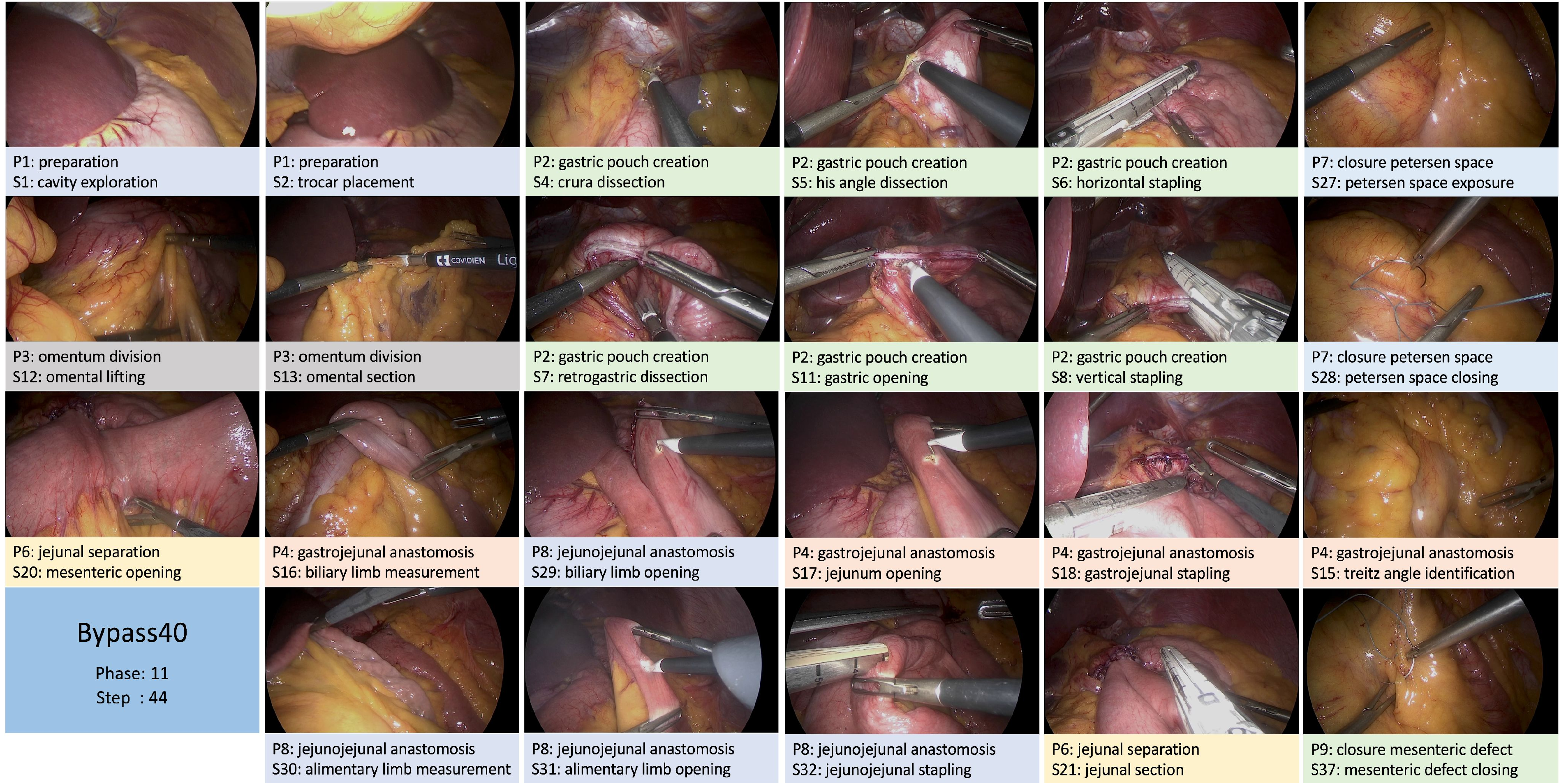}
\end{subfigure}
\par\smallskip
\begin{subfigure}[b]{1\textwidth}
    \centering
    \includegraphics[width=\textwidth]{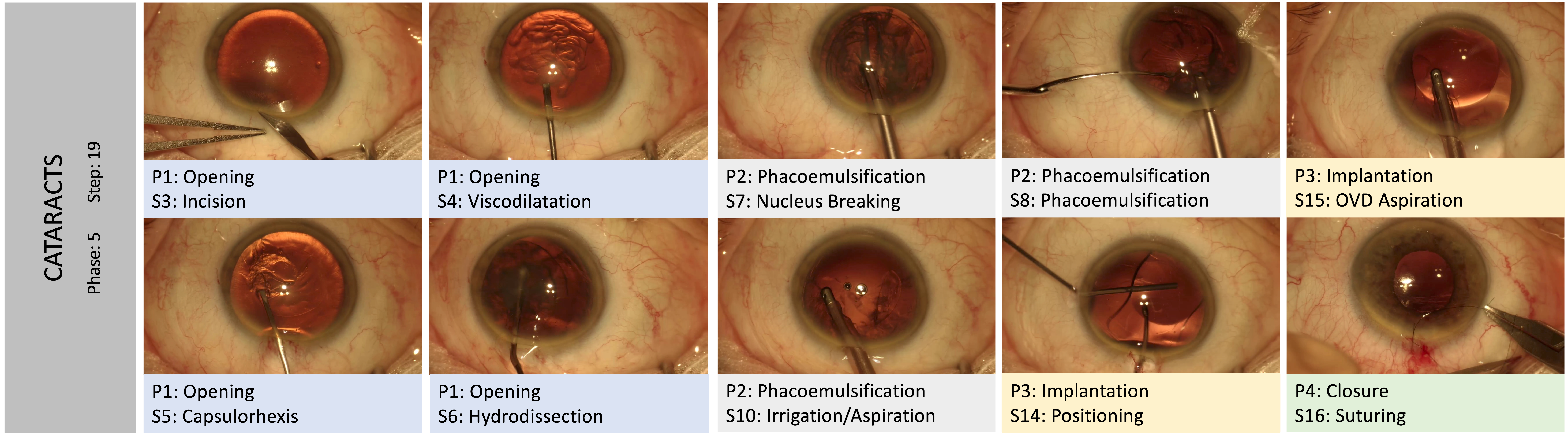}
\end{subfigure}
\caption{Sample images from Bypass40 and CATARACTS datasets. Each column of Bypass40 images present similar steps.} \label{fig1:sample_images}
\end{figure*}

To address this issue, a few works \cite{DBLP:journals/corr/BodenstedtWKMMK17, Funke2018, Yengera2018LessIM, DBLP:journals/corr/abs-1812-00033/YuTSModel2019} have proposed methods based on semi-supervision. These approaches involve either pre-training the model on proxy tasks or training on synthetic labels generated by a teacher model trained on a small subset for phase recognition. Unlike these works, inspired by \cite{Nwoye2019WeaklySC} and \cite{FuentesHurtado2019}, we address the annotation scarcity issue by proposing a weakly supervised learning approach utilizing relatively economical annotations.

The main contributions of our work are summarized as follows:
\begin{enumerate}
    \item We propose a weakly supervised learning method for surgical workflow analysis to tackle the problem of fine-grained surgical activity (step) recognition. We exploit the hierarchical step-phase relationships and utilize easier-to-annotate weak phase annotations on videos with missing step annotations. 
    \item We introduce a novel dependency loss to enforce the weak supervision and encode the step-phase hierarchical relationship as a matrix. By optimizing for this loss, it encourages the model to learn possible step sequences and transitions from videos with only phase annotations. 
    \item We present an end-to-end model consisting of ResNet-50 and Single-Stage Temporal Convolutional Network (SS-TCN) to learn both visual and temporal cues jointly.
    \item We extend the CATARACTS\footnote{https://cataracts2020.grand-challenge.org/} dataset (containing step annotations) with phase annotations. These annotations will be released upon acceptance of this manuscript.
    \item  We extensively evaluate our approach on two surgical video datasets, namely Bypass40 \cite{Ramesh2021} and CATARACTS \cite{AlHajj2019}, demonstrating the effectiveness and generalizability of our method. 
\end{enumerate}


\section{Related Work}

\subsection{Surgical Activity Recognition} Research on developing deep learning methods for surgical phase recognition has seen significant progress with initial works of EndoNet \cite{Twinanda2017} and DeepPhase \cite{Zisimopoulos2018DeepPhaseSP} on cholecystectomy and cataract surgeries, respectively. EndoNet proposed a Convolutional Neural Network (CNN) followed by a hierarchical Hidden Markov Model (HMM) to perform both phase and tool detection. Similarly, DeepPhase introduced an architecture with ResNet \cite{He2016} and Recurrent Neural Network (RNN), instead of HMMs, for temporal modeling, for both phase recognition and tool detection. EndoLSTM \cite{Twinanda2016SingleAM, Twinanda2017VisionbasedAF} extended EndoNet by utilizing a Long Short-Term Memory (LSTM) for temporal refinement of spatial features. Similarly, SV-RCNet \cite{Jin2018} trained a ResNet and LSTM model end-to-end and proposed a prior knowledge inference scheme for surgical phase recognition. MTRCNet-CL \cite{Jin2020MultiTaskRC} presented a multi-task model to detect tool presence and perform phase recognition along with a novel correlation loss to capture the relationship between tool presence and phase identification. Recently, TeCNO \cite{Czempiel2020TeCNOSP} adapted the multi-stage Temporal Convolutional Network (MS-TCN) \cite{farha2019mstcn} architecture for online surgical phase prediction by implementing causal convolutions \cite{wavenet2016}.

On the other hand, step recognition has seen a spark in research with the initial work of \cite{Charriere2014}. A Content-Based Video Retrieval (CBVR) system, for real-time step recognition, was proposed utilizing a novel pupil center and scale tracking method as pre-processing of motion features. In \cite{Charrire2017}, the CBVR system along with surgical tool presence information was used as input to statistical models
consisting of Bayesian Network and HMMs
for multi-level 
online recognition of step and phase. Recently, MTMS-TCN \cite{Ramesh2021} adapted TeCNO utilizing TCNs for multi-level
online recognition of step and phase.
In this work, we build upon the architectures of TeCNO and MTMS-TCN by utilizing a variant of MS-TCN in an end-to-end fashion for online step recognition. 

\subsection{Weak Supervision} \label{sec:related_weak} Weak supervision has seen a great interest in the medical computer vision community to tackle the need for high-volume annotated datasets that are difficult to generate. Some of the interesting applications of weak supervision are seen in surgical tool localization \cite{Nwoye2019WeaklySC}, tool segmentation \cite{FuentesHurtado2019}, cancerous tissue segmentation \cite{7971941}, and detection of the region of interest in chest X-rays and mammograms \cite{Hwang2016}. 
To reduce the number of labeled videos, most of the recent research works in phase recognition have proposed \textcolor{highlight}{approaches based} on semi-supervised learning. These approaches follow a similar strategy of pre-training the models on different proxy tasks of frame-sorting \cite{DBLP:journals/corr/BodenstedtWKMMK17}, predicting the temporal distance between multiple frames \cite{Funke2018}, and predicting the remaining surgery duration \cite{Yengera2018LessIM}. The most closely related work to this paper in terms of objectives is \cite{DBLP:journals/corr/abs-1812-00033/YuTSModel2019}, which proposed a teacher/student approach for phase recognition in scenarios of extreme manual annotation scarcity ($\le25\%$ of the training set). The teacher model (trained on a small set) generated synthetic phase annotations for a large number of videos on which the student model was then trained. 

Weakly supervised coarse-to-fine methods have received considerable interest in the computer vision community \cite{Taherkhani2019WeakFine, Touvron2020GrafitLF, Bukchin2020FinegrainedAC} for image classification. \cite{Taherkhani2019WeakFine} proposed an image-based weakly supervised end-to-end model for object classification consisting of a CNN followed by two self-expressive layers. One self-expressive layer \textcolor{highlight}{captures} the global structures through coarse labels and the other \textcolor{highlight}{captures} the local structures for fine-grained classification. \cite{Touvron2020GrafitLF} tackled the problem of learning finer representations from coarser labels without any fine-grained labels. Their proposed method consists of CNN based trunk-target network that learns coarse representations from labels and finer representations with nearest-neighbor classifier objective. Recently, \cite{Bukchin2020FinegrainedAC} tackled the problem of Coarse-to-Fine Few-Shot (C2FS) and proposed a novel `angular normalization' module that effectively combines supervised and self-supervised contrastive pre-training for C2FS.

Although these previous works in the vision community propose weakly supervised learning methods exploiting hierarchical structures, the focus solely lies on object recognition in natural images containing a single object in each image. In this work, we focus on weakly supervised learning from videos instead of images. We aim to recognize fine-grained activity, as opposed to object, exploiting the temporal information available in videos. In particular, we target fine-grained surgical activity recognition on videos from endoscopic procedures on two different types of surgeries, i.e., gastric bypass and cataract.


\section{Methodology}

\begin{figure*}[!t]
\centering
\includegraphics[width=0.9\textwidth]{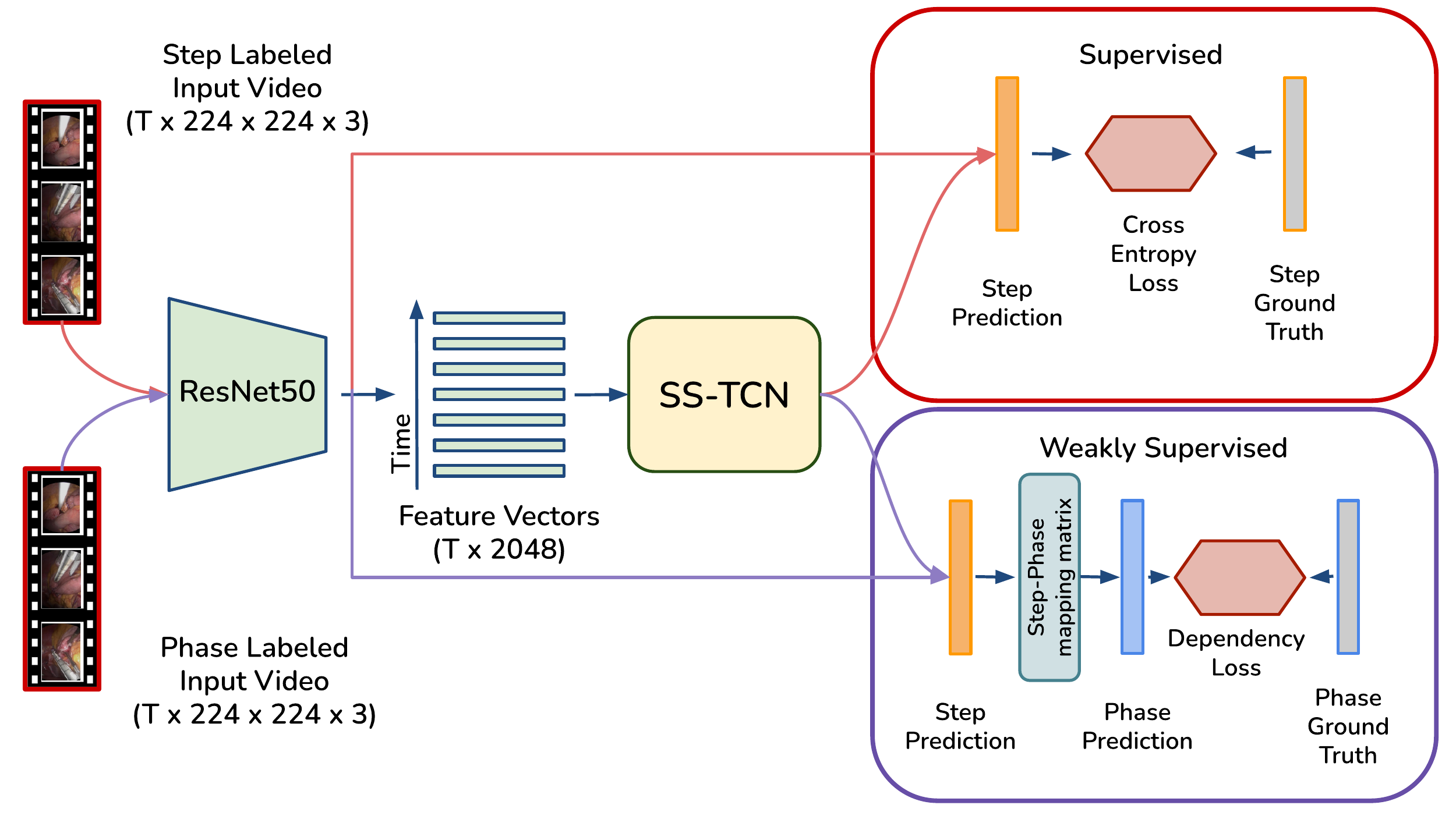}
\caption{Overview of our end-to-end Spatio-temporal model setup: ResNet50 + SS-TCN (Single-Stage Temporal Convolutional Networks). When step labels are available, the model is trained through the supervised pathway (red) and weakly supervised pathway (purple) utilizing phase labels. The model is trained end-to-end in a single learning stage.} \label{fig2:model_arch}
\end{figure*}

The overview of our proposed method is presented in Fig. \ref{fig2:model_arch}. In this section, we first present our end-to-end Spatio-temporal (ResNet-50 + SS-TCN) model for the task of fine-grained activity, i.e, step, recognition. Then we introduce the phase-step dependency loss for weak supervision of step recognition using phase annotation.

\subsection{Spatio-temporal Model} 

Our weakly supervised step recognition network consists of a ResNet-50 model for visual feature extraction followed by \textcolor{highlight}{an} SS-TCN for modeling the recognition problem temporally. The complete model is trained in an end-to-end fashion. The overview of the model setup is depicted in Fig. \ref{fig2:model_arch}. 

For phase segmentation, ResNet-50 \cite{He2016_V2} has been successfully employed as \textcolor{highlight}{the} backbone in many previous works \cite{DBLP:journals/corr/abs-1812-00033/YuTSModel2019, Czempiel2020TeCNOSP, Jin2018, Jin2020MultiTaskRC}. In this work, we utilize the same architecture for visual feature extraction. We use a single-stage TCN (SS-TCN), a single-stage variant of MS-TCN, to learn the spatial coherence across video frames. The choice of SS-TCN was motivated by the work of \cite{Ramesh2021} where MS-TCN did not provide \textcolor{highlight}{a} significant improvement over SS-TCN for both the step and phase recognition. Following the design of MS-TCN, the SS-TCN contains neither pooling layers nor fully connected layers and is constructed with only temporal convolutional layers, specifically dilated residual layers performing dilated convolutions. With the aim of online activity segmentation, we perform at each layer causal convolutions \textcolor{highlight}{\cite{wavenet2016,Czempiel2020TeCNOSP,Ramesh2021} that depend} only on the current frame and $n$ previous frames.

The complete model takes an input video consisting of $T$ frames $x_{1:T}$. The ResNet-50 maps $224\times224\times3$ RGB images to a feature space of size $N_f=2048$. These frame-wise features are collected over time and are inputs to the TCN model that predicts $\hat{y}^s_{1:T}$ where $\hat{y}^s_t$ is the class label for the current timestamp $t$, $t \in [1, T]$. Since step recognition is a multi-class classification problem that exhibits \textcolor{highlight}{an} imbalance in the class distribution, softmax activation and class-weighted cross-entropy loss are utilized. Additionally, the dependency loss used when step labels are not available also relies on softmax activation and weighted cross-entropy loss, utilizing phase labels instead. The class weights for both steps and phases are calculated using the median frequency balancing \cite{7410661} on the training set. The total loss is given by:
\begin{equation}
    \mathcal{L}_{total} = \delta_{step}\cdot\mathcal{L}_{step} + (1-\delta_{step})\cdot\mathcal{L}_{dep},
\end{equation}\label{eqn1:total_loss}where $\mathcal{L}_{step}$ represents weighted cross-entropy loss for steps, $\mathcal{L}_{dep}$ is the step-phase dependency loss (subsection \ref{dep_loss}), and $\delta_{step}$ is a binary variable that indicates if the video contains step labels.

\subsection{Weak Supervision: Step-Phase dependency loss}\label{dep_loss}

Steps and phases are two types of activities describing the surgical workflow that are defined at different levels of granularity and possess an inherent hierarchical relationship \cite{Katic2015, Ramesh2021}. Steps are defined at a higher level of detail compared to phases. This brings about lower inter-class variances between steps, compared to phases, making it a more complex task to clearly define and distinguish between them. The challenges can be seen in the sample images presented in Fig. \ref{fig1:sample_images}. \textcolor{highlight}{For instance,} in the Bypass40 dataset, similar actions are performed across different steps belonging to different phases. Dissection is performed in at least 7 steps spread across 3 different phases. Similarly, Stapling is performed in 5 steps across 4 different phases. 
Designing and training a deep learning model to distinguish between these similar steps poses a great challenge. Even the state-of-the-art method, MTMS-TCN \cite{Ramesh2021}, trained on a fully annotated dataset achieves an accuracy of $\sim$76\% with a precision of $\sim$56\%, accentuating the difficulty of the problem. The class imbalance further creates a challenge for training deep learning models that require large datasets with plenty of samples for each class. 

In the scenario presented in this paper where the number of annotations is scarce, the recognition difficulties increase drastically. To overcome some of the challenges, this work proposes a weakly supervised approach that utilizes labels of less granular activities, i.e., phases. Phase information alone could help the model in two ways. Firstly, phase information could help the model reduce errors related to recognizing similar looking steps, e.g., `S6: horizontal stapling' and `S18: gastrojejunal stapling', belonging to two different phases. Secondly, we can gather a smaller subset of probable steps that could occur in a given phase eliminating the rest. For example, given the phase to be `Phacoemulsification' of cataract surgery, only 5 out of 19 steps are likely to occur (Table \ref{tab1:cataracts_phst}). Similarly, a phase such as `P5: anastomosis test' in the Bypass40 dataset, reduces the possible steps to 7 out of 44 (Table \ref{tab2:lrygb_phst}). Here, the phase information provides cues to the model to learn to distinguish between steps belonging to the subset rather than the whole set.
Thus we hypothesize that the additional available weak phase information could be very beneficial for step recognition in the low data regime.


We propose to represent the relationship as a step-phase mapping matrix $M_{s \rightarrow p}$, where the elements $m_{ij}$ of the matrix are binary indicator variables which are $1$ if step $s_i$ occurs in phase $p_j$. The matrix encodes the weak information \textcolor{highlight}{about} which steps can occur in a particular phase and does not provide details of their occurrence, duration, and/or order. To enforce this weak link between steps and phases, the step predictions $\hat{y}^s_t$ of our Spatio-temporal model (as described earlier) are linearly transformed by $M_{s \rightarrow p}$ into the phase space. Then a weighted cross-entropy loss ($\mathcal{L}_{CE}$) captures the similarity between the phase labels ($y^p_t$) and the transformed predictions ($M_{s \rightarrow p} \times \hat{y}^s_t$) of the model. The dependency loss ($\mathcal{L}_{dep}$) is given by:
\begin{equation}
    \mathcal{L}_{dep} = \mathcal{L}_{CE}(y^p_t, M_{s \rightarrow p} \times \hat{y}^s_t).
\end{equation}

\begin{table*}
\caption{Phases and steps for the cataract procedure}\label{tab1:cataracts_phst}
\centering
\begin{tabular*}{0.9\textwidth}{c @{\extracolsep{\fill}} cccccc}
\hline\noalign{\smallskip}
\textbf{\textcolor{teal}{Phases}} & Idle &  Opening & Phacoemulsification & Implantation & Closure\\
\noalign{\smallskip}\hline\noalign{\smallskip}
\multirow{3}{*}{\textbf{\textcolor{teal}{Steps}}}   & Idle                     &      Idle &       Idle &    Idle &     Idle \\
     &            &      Toric Marking &       Nucleus Breaking &    Incision  &     Suturing \\
     &    &      Implant Ejection &       Phacoemulsification &    Viscodilatation &     Sealing Control \\
     & & Incision & Vitrectomy & Preparing Implant
& Wound Hydration \\
     & & Viscodilatation & Irrigation/Aspiration
 & Manual Aspiration
& 
\\
     & & Capsulorhexis & & Implantation & \\
     & & Hydrodissection & & Positioning & \\
     & &  & & OVD Aspiration & \\
\noalign{\smallskip}\hline
\end{tabular*}
\end{table*}

\begin{table*}
\caption{Phases and steps for the laparoscopic Rue-en-Y gastric bypass procedure}\label{tab2:lrygb_phst}
\centering
\begin{tabular*}{0.93\textwidth}{l L}
\hline\noalign{\smallskip}
\textbf{\textcolor{teal}{Phases}} & \textbf{\textcolor{teal}{Steps}}\\
\noalign{\smallskip}\hline\noalign{\smallskip}
P1: preparation &
S0: null step, 
S1: cavity exploration, 
S2: trocar placement, 
S3: retractor placement,
S14: adhesiolysis,
S22: gastric tube placement\\
\noalign{\smallskip}
P2: gastric pouch creation &
S0: null step, 
S4: crura dissection,
S5: his angle dissection,
S6: horizontal stapling,
S7: retrogastric dissection,
S8: vertical stapling,
S9: gastric remnant reinforcement,
S10: gastric pouch reinforcement,
S11: gastric opening,
S22: gastric tube placement,
S43: calibration\\
\noalign{\smallskip}
P3: omentum division & 
S0: null step, 
S12: omental lifting,
S13: omental section,
S14: adhesiolysis\\
\noalign{\smallskip}
P4: gastrojejunal anastomosis & 
S0: null step, 
S15: treitz angle identification,
S16: biliary limb measurement,
S17: jejunum opening,
S18: gastrojejunal stapling,
S19: gastrojejunal defect closing,
S26: gastrojejunal anastomosis reinforcement,
S30: alimentary limb measurement\\
\noalign{\smallskip}
P5: anastomosis test & 
S0: null step, 
S22: gastric tube placement,
S23: clamping,
S24: ink injection,
S25: visual assessment,
S26: gastrojejunal anastomosis reinforcement,
S39: coagulation\\
\noalign{\smallskip}
P6: jejunal separation & 
S0: null step, 
S20: mesenteric opening,
S21: jejunal section\\
\noalign{\smallskip}
P7: closure petersen space & 
S0: null step, 
S27: petersen space exposure,
S28: petersen space closing
\\
\noalign{\smallskip}
P8: jejunojejunal anastomosis & 
S0: null step, 
S29: biliary limb opening,
S30: alimentary limb measurement,
S31: alimentary limb opening,
S32: jejunojejunal stapling,
S33: jejunojejunal defect closing,
S34: jejunojejunal anastomosis reinforcement,
S35: staple line reinforcement
\\
\noalign{\smallskip}
P9: closure mesenteric defect & 
S0: null step, 
S36: mesenteric defect exposure,
S37: mesenteric defect closing,
S38: anastomosis fixation,
\\
\noalign{\smallskip}
P10: cleaning coagulation &
S0: null step, 
S39: coagulation,
S40: irrigation aspiration
\\
\noalign{\smallskip}
P11: disassembling & 
S0: null step, 
S40: irrigation aspiration,
S41: parietal closure,
S42: trocar removal
\\
\noalign{\smallskip}\hline
\end{tabular*}
\end{table*}

\section{Experimental Setup}

In this section, we discuss the experimental setup of our method. First, we present the datasets used for evaluation. Next, we discuss the experimental study followed by the training setup and evaluation metrics.  

\subsection{Datasets}

\subsubsection{Bypass40} The {\itshape Bypass40} dataset \cite{Ramesh2021} consists of 40 videos of LRYGB procedures with resolution $854 \times 480$ or $1920 \times 1080$ pixels recorded at 25 fps. Each frame is manually assigned to one of the 11 phases and one of the 44 steps \cite{Ramesh2021}. For example, steps such as {\itshape gastric opening, gastric tube placement, horizontal stapling}, and {\itshape vertical stapling} occur in {\itshape gastric pouch creation} phase. A detailed list of phases and steps along with their hierarchical relationship is presented in  Table \ref{tab2:lrygb_phst}.
For more information, we ask the readers to refer to \cite{Ramesh2021}. 
\textcolor{highlight}{
We split the 40 videos into 24, 6, and 10 videos for training, validation, and test sets, respectively, and sub-sampled them at 1 frame-per-second (fps). This amounts to 150k, 40k, and 65k images in each set.
}
The images are resized to ResNet-50's input dimension of $224 \times 224$, and the training dataset is augmented by applying horizontal flip, saturation, and rotation. 

\subsubsection{CATARACTS}

The CATARACTS dataset, proposed in \cite{AlHajj2019}, contains 50 videos of cataract surgery. With the recent CATARACTS2020 challenge, the dataset has been released with step annotations. Similar to \cite{Charrire2017}, we define a phase ontology for available step labels. 
Cataract surgery consists of 5 phases and 19 steps that are summarized in Table \ref{tab1:cataracts_phst}.
The dataset is extended with phase labels that is automatically generated using the available step annotations and the ontology presented in Table \ref{tab1:cataracts_phst}. For each frame in a video, the phase label is obtained by a simple lookup of the step label in Table \ref{tab1:cataracts_phst}. The only constraint while generating phase labels is when there are steps that can occur in several phases. In this case, the phase of the immediately preceding frame is assigned to the current frame. Since the only steps that occur in more than one phase are Idle, Incision, and Viscodilatation, and they do not occur at the beginning or at the end of a phase, it is therefore always possible to identify the correct phase by checking the phase of the previous step.
Since very few steps occur in multiple phases, the automatically generated phase labels by table lookup are accurate and do not require expert knowledge or verification from a clinical expert.

We split the 50 videos (following the challenge\footnote{https://www.synapse.org/\#!Synapse:syn21680292/wiki/601563}) into 25, 5, and 20 videos for training, validation, and test sets, respectively. 
\textcolor{highlight}{
Each set consists of 66k, 3.5k, and 11.8k frames extracted at 1 fps from the videos.
The frames are resized from $1920 \times 1080$ to $224 \times 224$, and the training set is augmented with horizontal flip, saturation, and rotation.
}

\subsection{Study} 
To demonstrate the effectiveness of our approach, we train and evaluate different configurations of the model. Given $n$ videos, of which $k$ are annotated with steps and the rest ($n-k$) are weakly annotated with phases, the Spatio-temporal model is trained in the proposed weakly supervised setting utilizing the dependency loss, presented as `DEP'. 
To analyze the efficacy of `DEP', we compare it against the Spatio-temporal model trained only on $k$ videos in a fully-supervised approach for the task of step recognition, which we refer to as `FSA'. 
Additionally, we add a state-of-the-art semi-supervised learning method proposed by Yu et al. \cite{Yu2018LearningFA} to our results. Yu et al. \cite{Yu2018LearningFA}, proposed a teacher/student semi-supervised learning method where both the teacher and student models consisted of spatial and temporal components, CNN-biLSTM-CRF and CNN-LSTM respectively. As noted in Section \ref{sec:related_weak}, \cite{Yu2018LearningFA} is a closely related work in the literature to the work presented in this paper.
Hence, we have implemented and adapted the method of Yu et al. \cite{Yu2018LearningFA} for the task of step recognition.
We repeat all the experiments for different values of $k\in \{3,6,12,18\}$. 

Furthermore, to analyze the influence of the number of additional videos with phase labels on the model performance, we conduct experiments where we fix $k$ videos with step annotations and vary the number of videos with phase annotations from $0$ to $n-k$ (i.e., 3, 6, 12, etc.).

\subsection{Training} The ResNet-50 model is initialized with weights pre-trained on ImageNet. The complete ResNet-50 + SS-TCN model is then trained end-to-end for the task of step recognition. Since SS-TCN models the temporal information in an online setup, features from all the past frames in the video \textcolor{highlight}{needs} to be cached. To achieve this, a feature buffer is maintained to store features from the spatial model of the past frames. The feature buffer is reset at the end of the video. In all the experiments, the model is trained for $50$ epochs with a learning rate of 1e-5, weight regularization of 5e-4, and a batch size of 64. The test results presented are from the best performing model on the validation set. 
The models were implemented in PyTorch and trained on NVIDIA RTX 2080 Ti.

\subsection{Evaluation Metrics} To effectively analyze our models, we observe the accuracy (ACC), precision (PR), recall (RE), and F1 score (F1) metrics used in related publications \cite{Czempiel2020TeCNOSP, Jin2018, Jin2020MultiTaskRC}. Accuracy quantifies the total correct classification of activity in the whole video. PR, RE, and F1 are computed class-wise, defined as:
\begin{equation}
    PR=\frac{| GT\cap P |}{| P |},\  RE=\frac{| GT\cap P |}{| GT |},\  F1= \frac{2}{\frac{1}{PR} + \frac{1}{RE}},
\end{equation}
where GT and P represent the ground truth and prediction for one class, respectively. These values are averaged across all the classes to obtain PR, RE, and F1 for each video in the test set. All four metrics, computed per video, are averaged across all the videos in the test set. Furthermore, where applicable, standard deviations are also computed across all the videos in the test set.


\section{Results and Discussions}

\subsection{Bypass40}

\subsubsection{Effect of weak supervision} To quantitatively evaluate our method, the results of step recognition on the test set are presented in  Table \ref{tab4:bypass40_results}. The table contains the results of our model with a varying number of videos in the training set labeled with steps (3, 6, 12, and 18) along with the rest of the training set containing phase annotations. The introduction of dependency loss `DEP' for weak supervision significantly improves the performance over the model (FSA) trained only on the step labeled subset of the dataset. 
We notice a 10-13\% improvement of the model trained with `DEP' loss containing only 3 videos annotated with steps. 
Similarly, we see a 10-13\% and 5-7\% increase in performance in all the metrics of the `DEP' model in experiments corresponding to 6 and 12 step annotated videos, respectively. Interestingly, our `DEP' model, trained on a dataset with 50\% of step and 50\% of phase annotated videos, achieves performance close to the upper baseline `FSA' model trained on the whole fully labeled dataset. 

\ifthenelse{\boolean{image_metrics}}{
\begin{table}
\caption{Bypass40: Effect of weak supervision on varying amount of step labeled videos. Accuracy (ACC), precision (PR), recall (RE), and F1-score (F1) (\%) are reported. `FSA' denotes the model trained for step recognition without any phase annotations. `DEP' denotes the dependency loss 
added for weak supervision using phase labels on the remaining videos.}\label{tab4:bypass40_results}
\centering
\begin{tabular*}{0.48\textwidth}{c| @{\extracolsep{\fill}} cc|cccc}
\hline\noalign{\smallskip}
& \multicolumn{2}{c|}{\# Videos} & & & &\\
Model & Step & Phase &  ACC & PR & RE & F1\\
\noalign{\smallskip}\hline\noalign{\smallskip}
FSA & 3 (12\%) & - &      44.61 &       26.25 &    22.12 &     20.30 \\
Yu et al. \cite{Yu2018LearningFA} & 3 (12\%) & - &      42.39 & 20.11 & 19.68 & 17.81 \\
DEP & 3 (12\%) & 21   &      {\bfseries 56.46} &       {\bfseries 33.66} &    {\bfseries 33.10} &     {\bfseries 30.76} \\
\noalign{\smallskip}\hline\noalign{\smallskip}
 FSA & 6 (25\%) & - &      59.32 &       37.51 &    34.39 &     33.03 \\
 Yu et al. \cite{Yu2018LearningFA} & 3 (12\%) & - &      62.04 & 39.63 & 36.9 & 35.38 \\
 DEP & 6 (25\%) & 18  &      {\bfseries 67.16} &       {\bfseries 45.30} &     {\bfseries 41.72} &     {\bfseries 39.32} \\
 \noalign{\smallskip}\hline\noalign{\smallskip}
 FSA & 12 (50\%) & - &      67.39 &       44.85 &    41.65 &     40.63 \\
 Yu et al. \cite{Yu2018LearningFA} & 12 (50\%) & - &      66.41 & 48.67 & 43.91 & 42.84 \\
 DEP & 12 (50\%) & 12  &      {\bfseries 72.51} &       {\bfseries 53.59} &    {\bfseries 47.69} &     {\bfseries 47.74} \\
\noalign{\smallskip}\hline\noalign{\smallskip}
 FSA & 18 (75\%) & - &      72.01 &       49.48 &    47.81 &     47.26 \\
 Yu et al. \cite{Yu2018LearningFA} & 12 (50\%) & - &      73.98 & 52.41 & 50.97 & 49.95 \\
 DEP & 18 (75\%) & 6  &      {\bfseries 72.88} &       {\bfseries 53.32} &    {\bfseries 49.67} &     {\bfseries 48.33} \\
\noalign{\smallskip}\hline\hline\noalign{\smallskip}
 FSA & 24 (100\%) & - &      75.13 &       53.25 &    49.26 &     49.23 \\
\noalign{\smallskip}\hline
\end{tabular*}
\end{table}
}{}

\ifthenelse{\boolean{video_metrics}}{
\begin{table*}[ht!]
\caption{Bypass40: Effect of weak supervision on varying amount of step labeled videos. Accuracy (ACC), precision (PR), recall (RE), and F1-score (F1) (\%) are reported. `FSA' denotes the model trained for step recognition without any phase annotations. `DEP' denotes the dependency loss 
added for weak supervision using phase labels on the remaining videos.}\label{tab4:bypass40_results}
\centering
\begin{tabular*}{0.80\textwidth}{c| @{\extracolsep{\fill}} cc|cccc}
\hline\noalign{\smallskip}
& \multicolumn{2}{c|}{\# Videos} & & & &\\
Model & Step & Phase &  ACC & PR & RE & F1\\
\noalign{\smallskip}\hline\noalign{\smallskip}
FSA & 3 (12\%) & - &      $ 45.02 \pm 9.96 $ & $ 26.62 \pm 5.32 $ & $ 21.87 \pm 4.70 $  & $ 19.44 \pm 5.31 $ \\
Yu et al. \cite{Yu2018LearningFA} & 3 (12\%) & - &      $ 43.27 \pm 11.8 $ & $ 23.63 \pm 4.41 $ & $ 23.91 \pm 5.71 $ & $ 19.77 \pm 4.89 $ \\
DEP & 3 (12\%) & 21   &      \textbf{57.20 $\pm$ 8.31}  & \textbf{ 33.44 $\pm$ 6.04 } & \textbf{ 33.16 $\pm$ 6.37 } & \textbf{ 29.38 $\pm$ 6.11 } \\
\noalign{\smallskip}\hline\noalign{\smallskip}
 FSA & 6 (25\%) & - &      $ 59.80 \pm 10.17 $ & $ 37.19 \pm 8.52 $ & $ 35.93 \pm 7.31 $ & $ 32.15 \pm 8.03 $ \\
 Yu et al. \cite{Yu2018LearningFA} & 6 (25\%) & - &      $ 62.55 \pm 10.09 $ & $ 40.63 \pm 7.85 $ & $ 43.71 \pm 8.35 $ & $ 37.68 \pm 8.54 $ \\
 DEP & 6 (25\%) & 18  &     \textbf{ 68.03 $\pm$ 9.04 } & \textbf{ 50.05 $\pm$ 6.82 } & \textbf{ 45.86 $\pm$ 6.46 } & \textbf{ 42.05 $\pm$ 7.44 } \\
 \noalign{\smallskip}\hline\noalign{\smallskip}
 FSA & 12 (50\%) & - &     $ 68.26 \pm 8.31 $ & $ 47.57 \pm 7.84 $ & $ 44.74 \pm 7.59 $ & $ 41.30 \pm 8.44 $ \\
 Yu et al. \cite{Yu2018LearningFA} & 12 (50\%) & - &      $ 67.89 \pm 11.04 $ & $ 46.26 \pm 9.97 $ & $ 50.11 \pm 8.20 $ & $ 43.41 \pm 10.33 $ \\
 DEP & 12 (50\%) & 12  &      \textbf{ 73.43 $\pm$ 8.43 } & \textbf{ 53.40 $\pm$ 7.43 }  & \textbf{ 51.19 $\pm$ 8.20 }  & \textbf{ 48.34 $\pm$ 8.85 } \\
\noalign{\smallskip}\hline\noalign{\smallskip}
 FSA & 18 (75\%) & - &      $ 72.82 \pm 6.76 $ & $ 50.60 \pm 7.90 $   & $ 48.98 \pm 8.33 $ & $ 46.08 \pm 8.61 $ \\
 Yu et al. \cite{Yu2018LearningFA} & 18 (75\%) & - &      $ 73.33 \pm 10.15 $ & \textbf{ 54.78 $\pm$ 11.05 } & \textbf{ 57.21 $\pm$ 8.51 } & \textbf{ 51.72 $\pm$ 10.59 } \\
 DEP & 18 (75\%) & 6  &      \textbf{ 73.88 $\pm$ 8.11 } & $ 54.33 \pm 6.38 $ & $ 51.79 \pm 7.10 $  & $ 48.62 \pm 7.49 $ \\
\noalign{\smallskip}\hline\hline\noalign{\smallskip}
 FSA & 24 (100\%) & - &      $ 76.12 \pm 7.39 $ & $ 54.23 \pm 8.24 $ & $ 50.94 \pm 7.53 $ & $ 48.17 \pm 8.02 $ \\
\noalign{\smallskip}\hline
\end{tabular*}
\end{table*}
}{}

Moreover, the results of Yu et al. \cite{Yu2018LearningFA} semi-supervised method are also presented in Table \ref{tab4:bypass40_results} for different step annotated videos (3, 6, 12, and 18) used to train both teacher and student model.
The student model's performance increases by 3-8\% over `FSA' in all the metrics for 6 videos with step annotations. Furthermore, an increase of 6\% and 2\% is noticed in recall and F1-score above `FSA' with 12 step annotated videos. However, the method falls short of our proposed `DEP' method. We notice a 10-15\%, 2-6\%, and 1-6\% increase in performance in all the metrics of the `DEP' model over Yu et al. with 3, 6 and 12 step annotated videos, respectively. Although both methods use 100\% of the training videos for the task of step recognition, Yu et al. aim at exploiting the knowledge learned by an offline teacher model to generate pseudo labels for additional videos without step annotations while `DEP' aims to use weak supervision through phase annotations. Hence, the method of Yu et al. is limited by the knowledge learned by the teacher model which uses only $k$ step annotated videos although it learns from both current and future frames. On the other hand, the superior performance of the `DEP' model indicates the additional cues present in phase annotated videos, although weak, is advantageous and that the proposed method effectively utilizes this information in the lower data settings. 

\ifthenelse{\boolean{image_metrics}}{
\begin{table}
\caption{Bypass40: Effect of the number of phase annotated videos for step recognition using `DEP' loss for weak supervision. Accuracy (ACC), precision (PR), recall (RE), and F1-score (F1) (\%) are reported for setups with 6, 12, and 24 videos fully annotated with steps. 
}\label{tab5:bypass40_results2}
\centering
\begin{tabular*}{0.48\textwidth}{c| @{\extracolsep{\fill}} cc|cccc}
\hline\noalign{\smallskip}
& \multicolumn{2}{c|}{\# Videos} & & & &\\
Model & Step & Phase &  ACC & PR & RE & F1\\
\noalign{\smallskip}\hline\noalign{\smallskip}
FSA & 6     & -     &      59.32 &       37.51 &    34.39 &     33.03 \\
DEP & 6     & 3     &      61.84 &       41.41 &    35.53 &     34.26 \\
DEP & 6     & 6     &      67.11 &       45.77 &    42.78 &     41.68 \\
DEP & 6     & 12    &      {\bfseries 67.34} &       45.82 &    42.88 &     41.20 \\
DEP & 6     & 18    &      66.85 &       {\bfseries 49.29} &    {\bfseries 46.14} &     {\bfseries 43.67} \\
\noalign{\smallskip}\hline\noalign{\smallskip}
FSA & 12     & -     &      67.39 &       44.85 &    41.65 &     40.63 \\
DEP & 12     & 3     &      72.28 &       48.18 &    46.57 &     45.44 \\
DEP & 12     & 6     &      71.58 &       49.39 &    {\bfseries 48.32} &     46.26 \\
DEP & 12     & 12    &      {\bfseries 72.51} &       {\bfseries 53.59} &    47.69 &     {\bfseries 47.74} \\
\noalign{\smallskip}\hline\hline\noalign{\smallskip}
FSA  & 24       & -     &    75.13 &       53.25 &    49.26 &     49.23 \\
\noalign{\smallskip}\hline
\end{tabular*}
\end{table}
}{}

\ifthenelse{\boolean{video_metrics}}{
\begin{table}[t!]
\caption{Bypass40: Effect of the number of phase annotated videos for step recognition using `DEP' loss for weak supervision. Accuracy (ACC), precision (PR), recall (RE), and F1-score (F1) (\%) are reported for setups with 6, 12, and 24 videos fully annotated with steps. 
}\label{tab5:bypass40_results2}
\centering
\begin{tabular*}{0.48\textwidth}{c| @{\extracolsep{\fill}} cc|cccc}
\hline\noalign{\smallskip}
& \multicolumn{2}{c|}{\# Videos} & & & &\\
Model & Step & Phase &  ACC & PR & RE & F1\\
\noalign{\smallskip}\hline\noalign{\smallskip}
FSA & 6     & -     &      59.80 &       37.19 &    35.93 &     32.15 \\
DEP & 6     & 3     &      62.15 &       40.48 &    37.15 &     33.48 \\
DEP & 6     & 6     &      67.94 &       46.17 &    42.61 &     39.67 \\
DEP & 6     & 12    &      68.07 &       47.18 &    43.18 &     40.42 \\
DEP & 6     & 18    &      68.03 &       50.05 &    45.86 &     42.05 \\
\noalign{\smallskip}\hline\noalign{\smallskip}
FSA & 12     & -     &      68.26 &       47.57 &    44.74 &     41.30 \\
DEP & 12     & 3     &      72.79 &       50.10 &    48.39 &     45.06 \\
DEP & 12     & 6     &      72.43 &       53.02 &    51.20 &     47.26 \\
DEP & 12     & 12    &      73.43 &       53.40 &    51.19 &     48.34 \\
\noalign{\smallskip}\hline\hline\noalign{\smallskip}
FSA  & 24       & -     &    76.12  & 54.23  & 50.94  & 48.17  \\
\noalign{\smallskip}\hline
\end{tabular*}
\end{table}
}{}

\subsubsection{Effect of the amount of phase annotated videos}\label{sec:bypass40_r2} In  Table \ref{tab5:bypass40_results2}, we present the results of our model with a varying number of phase annotated videos. Utilizing 6 videos containing step annotations, the addition of phase labeled videos as weak supervision improves all metrics: accuracy, F1, precision, and recall. 
With 6 videos annotated with phases, the model performance increases by 7-8\% in all metrics over the baseline `FSA' model. The addition of more videos does not affect the accuracy but further improves both precision and recall by 4\%. This is due to our weakly-supervised method, which only provides supervision information if a step can occur in the given phase. This information helps to distinguish steps belonging to different phases, as opposed to steps belonging to the same phase. Therefore, the precision and recall of the model improve with more phase annotated videos, and no significant improvement in accuracy is seen. We see a similar trend when using 12 videos annotated with steps and increasing the number of videos annotated with phase labels. Thus, ultimately it is beneficial to train our method utilizing all additional videos in the dataset with phase annotations for weak supervision.

\ifthenelse{\boolean{image_metrics}}{
\begin{table}
\caption{CATARACTS: Effect of weak supervision on varying amount of step labeled videos.
Accuracy (ACC), precision (PR), recall (RE), and F1-score (F1) (\%) are reported. `FSA' denotes the model trained for step recognition without any phase annotations. `DEP' denotes the dependency loss added for weak supervision using phase labels on the remaining videos.}\label{tab6:cataracts_results1}
\centering
\begin{tabular*}{0.48\textwidth}{c| @{\extracolsep{\fill}} cc|cccc}
\hline\noalign{\smallskip}
& \multicolumn{2}{c|}{\# Videos} & & & &\\
Model & Step & Phase &  ACC & PR & RE & F1\\
\noalign{\smallskip}\hline\noalign{\smallskip}
    FSA & 3 (12\%) & -          &      43.55 &       35.9  &    23.63 &     24.67 \\
    Yu et al. \cite{Yu2018LearningFA} & 3 (12\%) & 22 & 55.02 & 41.9 & 35.68 &  36.52\\ 
    DEP & 3 (12\%) & 22         &      {\bfseries 61.54} &       {\bfseries 42.88} &    {\bfseries 40.38} &     {\bfseries 38.80} \\
\noalign{\smallskip}\hline\noalign{\smallskip}
    FSA & 6 (25\%) & -          &      64.48 &       47.87 &    37.19 &     40.08 \\
    Yu et al. \cite{Yu2018LearningFA} & 3 (12\%) & 22 & 71.76 & 50.21 & 46.54 & 47.24\\ 
    DEP & 6 (25\%) & 19         &      {\bfseries 70.30} &       {\bfseries 46.60} &    {\bfseries 46.50} &     {\bfseries 45.32} \\
 \noalign{\smallskip}\hline\noalign{\smallskip}
    FSA & 12 (50\%) & -         &      74.62 &       59.19 &    47.49 &     50.6 \\
    Yu et al. \cite{Yu2018LearningFA} & 12 (50\%) & 22 & 72.61 & 60.96 & 47.34 & 50.31\\ 
    DEP & 12 (50\%) & 13        &      {\bfseries 76.19} &       {\bfseries 62.10} &    {\bfseries 49.42} &     {\bfseries 51.14} \\
\noalign{\smallskip}\hline\noalign{\smallskip}
    FSA & 18 (75\%) & -        &      {\bfseries 78.94} &       {\bfseries 63.62} &    54.86 &     {\bfseries 57.22} \\
    Yu et al. \cite{Yu2018LearningFA} & 18 (75\%) & 22 & 73.67 & 60.39 & 55.06 & 54.46\\ 
    DEP & 18 (75\%) & 7        &      78.23 &       61.27 &    {\bfseries 56.72} &     56.73 \\
\noalign{\smallskip}\hline\hline\noalign{\smallskip}
    FSA & 25 (100\%) & -       &      79.23 &       67.33 &    54.39 &     56.61 \\
\noalign{\smallskip}\hline
\end{tabular*}
\end{table}
}{}

\ifthenelse{\boolean{video_metrics}}{
\begin{table*}[ht!]
\caption{CATARACTS: Effect of weak supervision on varying amount of step labeled videos.
Accuracy (ACC), precision (PR), recall (RE), and F1-score (F1) (\%) are reported. `FSA' denotes the model trained for step recognition without any phase annotations. `DEP' denotes the dependency loss added for weak supervision using phase labels on the remaining videos.}\label{tab6:cataracts_results1}
\centering
\begin{tabular*}{0.80\textwidth}{c| @{\extracolsep{\fill}} cc|cccc}
\hline\noalign{\smallskip}
& \multicolumn{2}{c|}{\# Videos} & & & &\\
Model & Step & Phase &  ACC & PR & RE & F1\\
\noalign{\smallskip}\hline\noalign{\smallskip}
FSA & 3 (12\%) & - &      $ 48.47 \pm 10.62 $ & $ 51.32 \pm 11.91 $ & $ 37.44 \pm 9.85 $  & $ 37.12 \pm 10.15 $ \\
Yu et al. \cite{Yu2018LearningFA} & 3 (12\%) & - &      $ 59.61 \pm 10.67 $ & $ 56.02 \pm 14.31 $ & \textbf{ 61.82 $\pm$ 14.45 } & $ 53.26 \pm 13.61 $ \\
DEP & 3 (12\%) & 22   &      \textbf{ 66.78 $\pm$ 12.21 } & \textbf{ 64.29 $\pm$ 12.50 }  & $ 59.73 \pm 11.93 $ & \textbf{ 58.31 $\pm$ 12.73 } \\
\noalign{\smallskip}\hline\noalign{\smallskip}
 FSA & 6 (25\%) & - &      $ 69.51 \pm 11.16 $ & $ 71.05 \pm 14.13 $ & $ 56.70 \pm 12.67 $  & $ 59.28 \pm 13.50 $   \\
 Yu et al. \cite{Yu2018LearningFA} & 6 (25\%) & - &      $ 74.62 \pm 8.22 $ & $ 67.71 \pm 11.48 $ & \textbf{ 75.93 $\pm$ 12.48 } & $ 67.67 \pm 12.46 $ \\
 DEP & 6 (25\%) & 19  &      \textbf{ 75.28 $\pm$ 11.50 }  & \textbf{ 71.84 $\pm$ 14.30 }  & $ 69.19 \pm 12.72 $ & \textbf{ 68.09 $\pm$ 13.97 } \\
 \noalign{\smallskip}\hline\noalign{\smallskip}
 FSA & 12 (50\%) & - &      $ 78.02 \pm 9.05 $  & $ 79.02 \pm 13.20 $  & $ 69.55 \pm 12.04 $ & $ 71.18 \pm 13.04 $  \\
 Yu et al. \cite{Yu2018LearningFA} & 12 (50\%) & - &      $ 77.84 \pm 12.55 $ & $ 71.48 \pm 13.41 $ & \textbf{ 79.92 $\pm$ 15.28 } & $ 72.96 \pm 14.46 $ \\
 DEP & 12 (50\%) & 13  &      \textbf{ 79.94 $\pm$ 9.17 } & \textbf{ 80.52 $\pm$ 12.93 } & $ 72.62 \pm 11.91 $ & \textbf{ 73.52 $\pm$ 13.29 } \\
\noalign{\smallskip}\hline\noalign{\smallskip}
 FSA & 18 (75\%) & - &      $ 82.5 \pm 8.07 $   & \textbf{ 82.58 $\pm$ 11.91 } & $ 76.05 \pm 11.62 $ & $ 77.39 \pm 12.12 $ \\
 Yu et al. \cite{Yu2018LearningFA} & 18 (75\%) & - &      $ 78.59 \pm 10.71 $ & $ 74.55 \pm 14.17 $ & \textbf{ 78.16 $\pm$ 12.64 } & $ 73.55 \pm 13.67 $ \\
 DEP & 18 (75\%) & 7  &     \textbf{ 82.64 $\pm$ 9.72 } & $ 82.20 \pm 13.70 $   & $ 77.32 \pm 12.70 $  & \textbf{ 77.67 $\pm$ 13.56 } \\
\noalign{\smallskip}\hline\hline\noalign{\smallskip}
 FSA & 25 (100\%) & - &      $ 83.37 \pm 9.50 $   & $ 85.29 \pm 12.05 $ & $ 78.96 \pm 11.93 $ & $ 80.09 \pm 13.34 $ \\
\noalign{\smallskip}\hline
\end{tabular*}
\end{table*}
}{}

\subsection{Cataracts}

\subsubsection{Effect of weak supervision} We quantitatively evaluate our method and present the results of step recognition in  Table \ref{tab6:cataracts_results1}. The table contains the results of our model, on a similar set of experiments as with {\itshape Bypass40}, by varying the number of videos in the training set labeled with steps (3, 6, 12, and 18) along with the rest of the training set containing phase annotations. We see a similar trend as with bypass where the `DEP' model outperforms `FSA'. 
We notice a 13-22\% improvement `DEP' model considering only 3 step annotated videos. 
Furthermore, we see a 6-13\% and 1-3\% increase in performance in all the metrics of the `DEP' model in experiments corresponding to 6 and 12 step annotated videos, respectively. We see that our method achieves a similar performance improvement on a relatively easier surgical workflow, such as cataracts, consistently surpassing the FSA in all labeled ratios. 
The semi-supervised method of Yu et al. achieves performance improvement of 16\%, 8\%, and 1.5\% over `FSA' in F1-score for experiments corresponding to 3, 6, and 12 videos, respectively. However, as seen earlier, it falls short of `DEP' by 5\%, 0.5\%, and 0.5\% in the F1-score for experiments corresponding to 3, 6, and 12 videos. 
Interestingly, Yu et al. achieves high recall on both datasets (Table \ref{tab4:bypass40_results} \& \ref{tab6:cataracts_results1}). On CATARACTS, it even outperforms the `DEP' model in recall in all the experiments but falls short significantly in precision. This could be credited to the student model which learns from imperfect pseudo labels generated by the teacher model. Since our proposed `DEP' model learns from true phase labels on additional videos its performance increases in both precision and recall. 
This validates the applicability of our approach to different surgical workflows.

\subsubsection{Effect of the amount of phase annotated videos} \label{sec:cataracts_r2} We present the results of our experiments, with a varying number of phase annotated videos, on CATARACTS in  Table \ref{tab7:cataracts_results2}. We notice that utilizing 6 step annotated videos with additional phase labeled videos improves all the metrics by 6-13\%. In particular, with 6 videos annotated with phases, we see a performance increase of 5\% in accuracy and  F1-score and 8\% in recall of the `DEP' model over the baseline `FSA'. The addition of more videos provides a fractional improvement in accuracy but further improves both recall and F1-score by 1-4\%. We see a similar trend when using 12 videos with step annotations reaffirming our hypothesis that it is beneficial to train our method utilizing all additional videos in the dataset with phase annotations for weak supervision.

\ifthenelse{\boolean{image_metrics}}{
\begin{table}
\caption{CATARACTS: Effect of the number of phase annotated videos for step recognition using `DEP' loss for weak supervision. Accuracy (ACC), precision (PR), recall (RE), and F1-score (F1) (\%) are reported for setups with 6, 12, and 25 videos fully annotated with steps. 
}\label{tab7:cataracts_results2}
\centering
\begin{tabular*}{0.48\textwidth}{c| @{\extracolsep{\fill}} cc|cccc}
\hline\noalign{\smallskip}
& \multicolumn{2}{c|}{\# Videos} & & & &\\
Model & Step & Phase &  ACC & PR & RE & F1\\
\noalign{\smallskip}\hline\noalign{\smallskip}
FSA & 6     & -     &      64.48 &       47.87 &    37.19 &     40.08 \\
DEP & 6     & 3     &      66.19 &       45.18 &    41.43 &     42.06 \\
DEP & 6     & 6     &      70.01 &       48.11 &    43.05 &     44.33 \\
DEP & 6     & 12    &      69.93 &       {\bfseries 48.30} &    43.70 &     45.01 \\
DEP & 6     & 19    &      {\bfseries 70.30} &      46.60 &    {\bfseries 46.50} &     {\bfseries 45.32} \\
\noalign{\smallskip}\hline\noalign{\smallskip}
FSA & 12     & -     &      74.62 &       59.19 &    47.49 &     50.6 \\
DEP & 12     & 3     &      74.29 &       54.99 &    45.60  &     48.48 \\
DEP & 12     & 6     &      {\bfseries 76.48} &       56.69 &    48.11 &     50.79 \\
DEP & 12     & 13    &      76.16 &       {\bfseries 61.54} &    {\bfseries 48.37} &     {\bfseries 51.27}  \\
\noalign{\smallskip}\hline\hline\noalign{\smallskip}
FSA & 25     & -     &    79.23 &       67.33 &    54.39 &     56.61 \\
\noalign{\smallskip}\hline
\end{tabular*}
\end{table}
}{}

\ifthenelse{\boolean{video_metrics}}{
\begin{table}[t!]
\caption{CATARACTS: Effect of the number of phase annotated videos for step recognition using `DEP' loss for weak supervision. Accuracy (ACC), precision (PR), recall (RE), and F1-score (F1) (\%) are reported for setups with 6, 12, and 25 videos fully annotated with steps. 
}\label{tab7:cataracts_results2}
\centering
\begin{tabular*}{0.48\textwidth}{c| @{\extracolsep{\fill}} cc|cccc}
\hline\noalign{\smallskip}
& \multicolumn{2}{c|}{\# Videos} & & & &\\
Model & Step & Phase &  ACC & PR & RE & F1\\
\noalign{\smallskip}\hline\noalign{\smallskip}
FSA & 6     & -     &      69.51 &       71.05 &    56.70 &     59.28 \\
DEP & 6     & 3     &      71.34 &       67.84 &    62.27 &     62.01 \\
DEP & 6     & 6     &      74.30 &       71.70 &    64.18 &     64.96 \\
DEP & 6     & 12    &      73.57 &       70.88 &    65.68 &     66.03 \\
DEP & 6     & 19    &      75.28 &       71.84 &    69.19 &     68.09 \\
\noalign{\smallskip}\hline\noalign{\smallskip}
FSA & 12     & -     &      78.02 &       79.02 &    69.55 &     71.18 \\
DEP & 12     & 3     &      77.60 &       78.26 &    68.60 &     69.87 \\
DEP & 12     & 6     &      80.11 &       81.60 &    72.46 &     73.98 \\
DEP & 12     & 13    &      79.94 &       80.52 &    72.62 &     73.52 \\
\noalign{\smallskip}\hline\hline\noalign{\smallskip}
FSA & 25     & -     &    83.37 &       85.29 &    78.96 &     80.09 \\
\noalign{\smallskip}\hline
\end{tabular*}
\end{table}
}{}

\subsection{Weak supervision on step predictions} 

\begin{figure*}[ht!]
     \centering
     \begin{subfigure}[b]{0.8\textwidth}
         \centering
         \includegraphics[width=\textwidth]{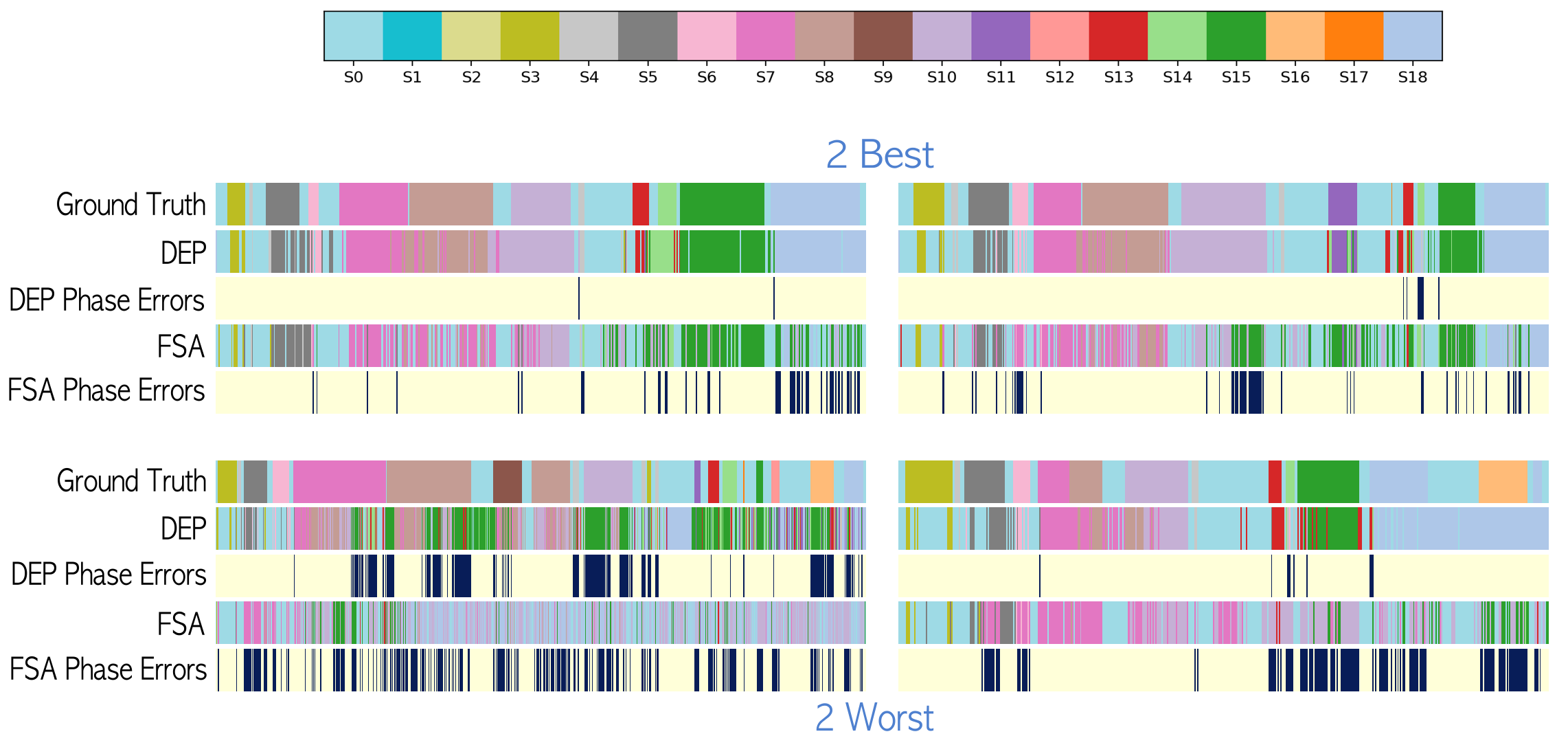}
         \caption{FSA vs DEP: 3 videos with step annotations.}
         \label{fig3a:3vid}
         \vspace*{4mm}
     \end{subfigure}
     \hfill
     \begin{subfigure}[b]{0.8\textwidth}
         \centering
         \includegraphics[width=\textwidth]{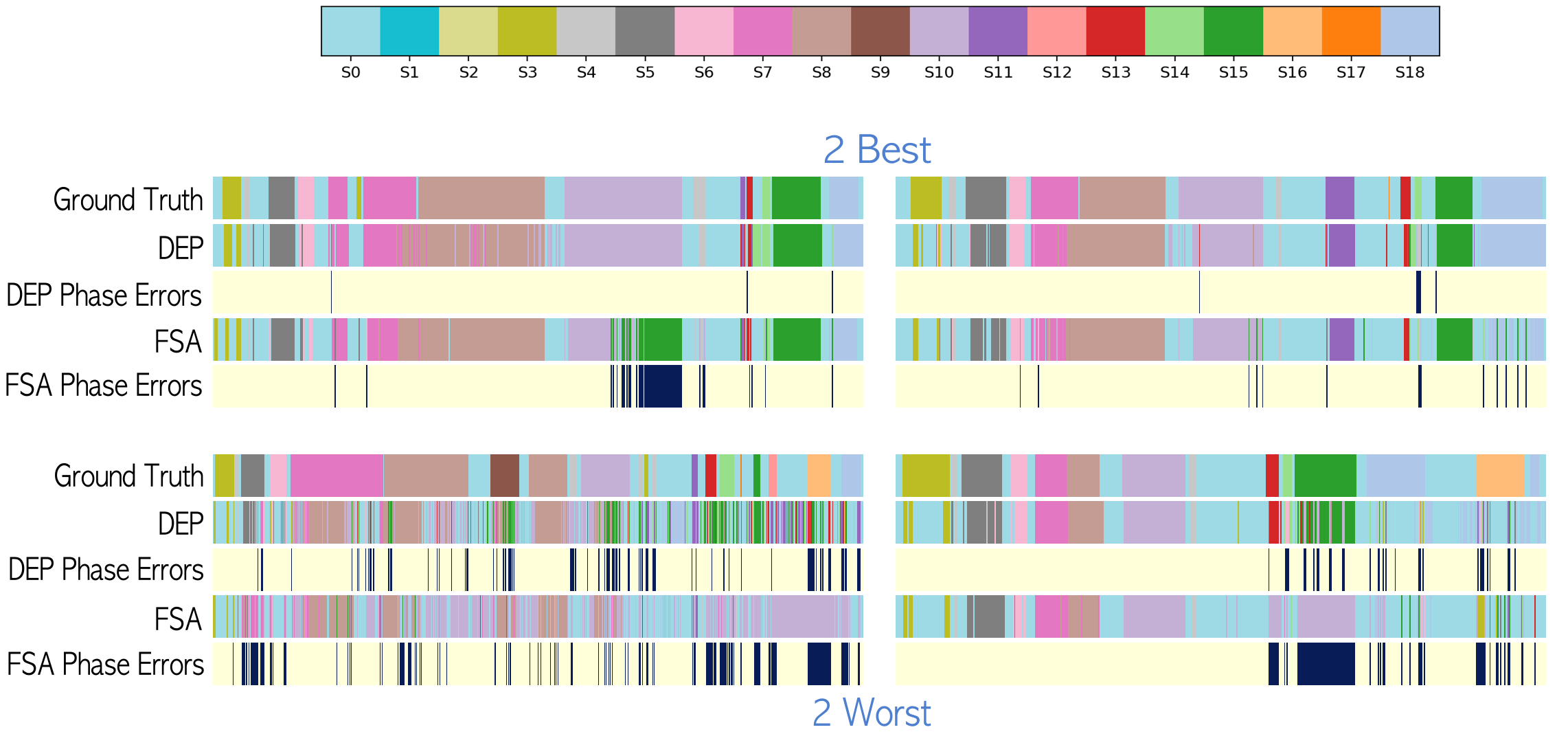}
         \caption{FSA vs DEP: 6 videos with step annotations.}
         \label{fig3b:6vid}
         \vspace*{4mm}
     \end{subfigure}
     \hfill
     \begin{subfigure}[b]{0.8\textwidth}
         \centering
         \includegraphics[width=\textwidth]{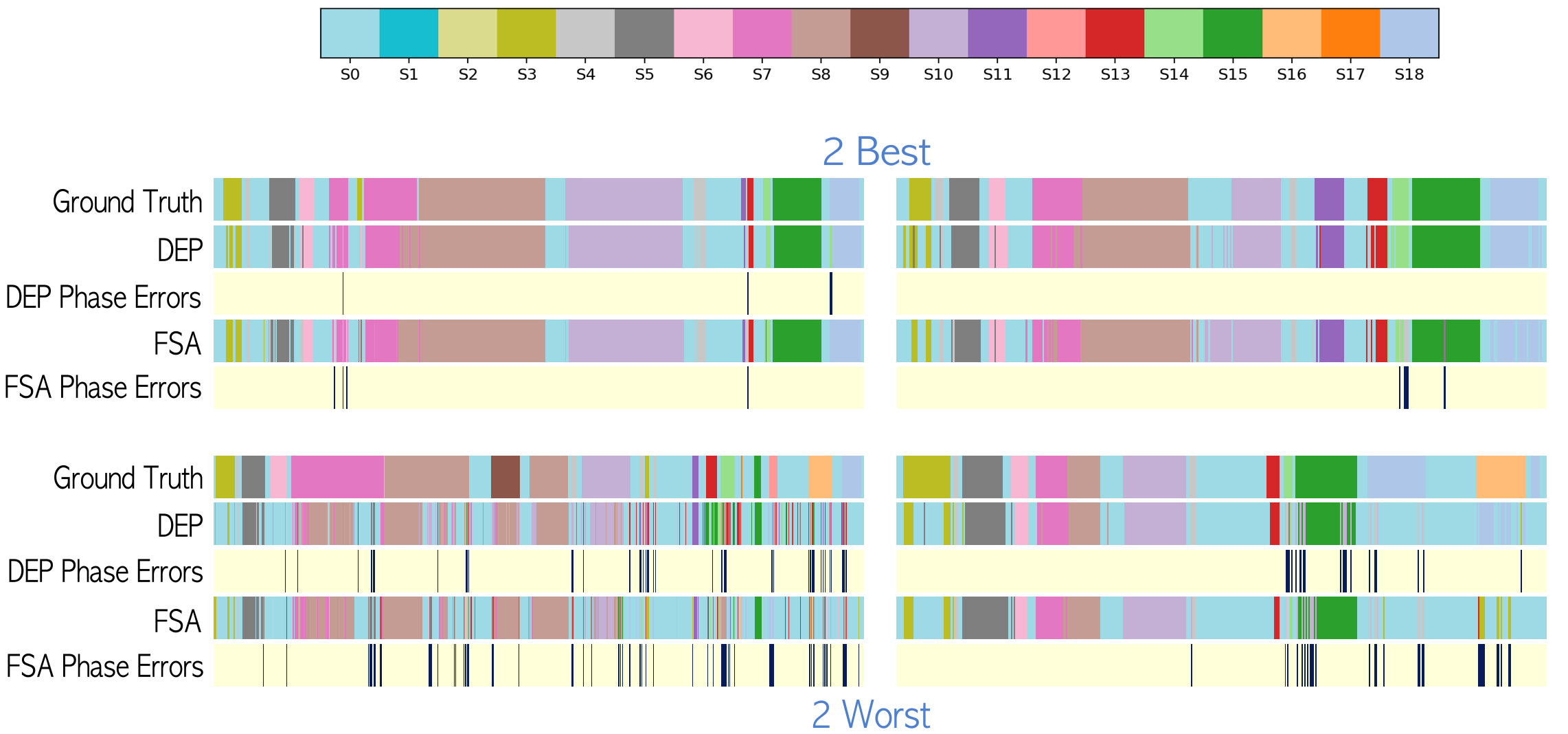}
         \caption{FSA vs DEP: 12 videos with step annotations.}
         \label{fig3c:12vid}
     \end{subfigure}
        \caption{Step predictions on two best and two worst videos on the CATARACTS dataset for different labeled ratios. For each video, we visualize the step prediction of ground truth, DEP model predictions, DEP model phase prediction errors, FSA model predictions, and phase prediction errors of FSA model.}
        \label{fig3:predictions}
\end{figure*}

To visualize the effectiveness of our method, we visualize the step predictions of our method on the CATARACTS dataset which contains fewer phases and steps thereby enabling us to render a simple and clearer graphical diagram. 
We compare the step predictions of our `DEP' model against `FSA' for 2 best and 2 worst videos in CATARACTS in Fig. \ref{fig3:predictions} for different labeled ratios (3, 6, and 12 videos with step annotations). Along with the step predictions we present the errors in the phase predictions for both models. The phase prediction error plot is computed as the errors in phase predictions derived from step predictions, using the step-phase mapping matrix, against ground truth phase predictions. Fig. \ref{fig3:predictions} clearly depicts the effectiveness of our method for different labeled ratios. By correcting for the phase labels through dependency loss, our `DEP' model is able to correct for corresponding step labels without explicit supervision for step recognition (e.g. S10, S15, S18). The top row of Fig. \ref{fig3a:3vid} shows this effect where we see a marked improvement in recognition of steps S18 (first video) and S10 (second video) by correcting for phase errors. 

\subsection{Limitations} In some cases, for example, S16 (Fig. \ref{fig3a:3vid}, \ref{fig3b:6vid}, \ref{fig3c:12vid}), correcting for phase errors does not improve step recognition. The step is misrecognized with another step that occurs in the same phase. This is an expected outcome due to the intrinsic limitations of our weakly supervised method using coarser phase labels. Given the phase to be `P2: gastric pouch creation' (Table \ref{tab2:lrygb_phst}), it is impossible for a model to differentiate between `crura dissection' and `his angle dissection' or between `horizontal stapling' and `vertical stapling'. As can be seen in Fig. \ref{fig1:sample_images}, the steps are quite similar in appearance and perform similar actions on the same anatomy (i.e., stomach or small intestine). This makes it challenging for a model to learn even when all the annotations are available. Furthermore, the phase information is too weak and does not provide any cues to better distinguish between the steps because both are valid steps in the current phase. Another limitation of our method is that adding more videos with phase annotations is not always beneficial. This limitation also stems from weak phase signals. If the fully supervised `FSA' model learns to separate steps belonging to different phases, i.e., it has no or few phase-step correspondence errors, then additional videos with phase labels add no significant value as the model, during training, makes no/few errors in phase-step correspondence that helps improve feature learning. The significant errors by the model would be the inter-class separation of steps belonging to the same phase. Learning good representations to reduce these errors without supervision is a challenging task that needs to be tackled in future works.

\textcolor{highlight}{Meanwhile, the effect of utilizing more phase annotated videos as weak supervision for improving the model performance on step recognition is presented in Tables \ref{tab5:bypass40_results2} \& \ref{tab7:cataracts_results2}. As observed in Sections \ref{sec:bypass40_r2} \& \ref{sec:cataracts_r2}, it is beneficial to train the `DEP' model utilizing all the additional phase annotated videos in the dataset for weak supervision. We also observe that in the lower data setting (6 videos with step annotations) model performance improves even when the phase annotated videos are increased from 12 to 18 (19 for cataracts). However, our study doesn't provide insights as to how many phase annotated videos are truly required to achieve the best performance by our proposed `DEP' model. This is another limitation of our study, irrespective of the complexity of the procedure, that is hindered by the size of the available labeled datasets (24 in Bypass40 \& 25 in CATARACTS). Understanding the extent of the `DEP' model would require extending these datasets which is an important direction that needs to be pursued in future studies.}

\section{Conclusion}

In this paper, we introduce a weakly-supervised learning method for surgical step recognition utilizing less demanding phase annotations. To model the weak supervision between steps and phases, we introduce a step-phase dependency loss and train a ResNet-50 + SS-TCN model end-to-end. The proposed method is extensively evaluated on a {\itshape Bypass40} dataset consisting of 40 LRYGB procedures and on the CATARACTS dataset containing 50 cataracts surgeries. The proposed `DEP' model significantly improves the step recognition metrics over the baseline `FSA' model for all the amounts of step annotations available. We hope that this work will inspire and foster future research in weak supervision for surgical workflow analysis utilizing multi-level descriptions of the workflow.
\\\\
{
{\bfseries Ethical approval} The surgical videos were recorded and anonymized following the informed consent of patients in compliance with the local Institutional Review Board (IRB) requirements. \\\\
{\bfseries Informed Consent} The patients consented to data recording.
}


\bibliographystyle{IEEEtran}
\bibliography{ref}

\end{document}